# Real-Time Human Action Recognition Using Locally Aggregated Kinematic-Guided Skeletonlet and Supervised Hashing-by-Analysis Model

Bin Sun, Shaofan Wang🄳, *Member, IEEE*, Dehui Kong🄳, *Member, IEEE*, Lichun Wang, and Baocai Yin

*Abstract*—3-D action recognition is referred to as the classification of action sequences which consist of 3-D skeleton joints. While many research works are devoted to 3-D action recognition, it mainly suffers from three problems: 1) highly complicated articulation; 2) a great amount of noise; and 3) low implementation efficiency. To tackle all these problems, we propose a real-time 3-D action-recognition framework by integrating the locally aggregated kinematic-guided skeletonlet (LAKS) with a supervised hashing-by-analysis (SHA) model. We first define the skeletonlet as a few combinations of joint offsets grouped in terms of the kinematic principle and then represent an action sequence using LAKS, which consists of a denoising phase and a locally aggregating phase. The denoising phase detects the noisy action data and adjusts it by replacing all the features within it with the features of the corresponding previous frame, while the locally aggregating phase sums the difference between an offset feature of the skeletonlet and its cluster center together over all the offset features of the sequence. Finally, the SHA model combines sparse representation with a hashing model, aiming at promoting the recognition accuracy while maintaining high efficiency. Experimental results on `MSRAction3D`, `UTKinectAction3D`, and `Florence3DAction` datasets demonstrate that the proposed method outperforms state-of-the-art methods in both recognition accuracy and implementation efficiency.

## I. INTRODUCTION

HUMAN action recognition is a very important research issue in recent years, which has wide applications in many research topics, such as human–computer interaction,

Manuscript received March 20, 2018; revised July 19, 2020 and June 16, 2021; accepted July 22, 2021. This work was supported in part by the National Natural Science Foundation of China under Grant 61772049, Grant 61632006, Grant U19B2039, Grant U1811463, and Grant 61876012; and in part by the Beijing Natural Science Foundation under Grant 4202003. This article was recommended by Associate Editor S. Ventura. *(Corresponding author: Dehui Kong.)*

Bin Sun is with the Beijing Institute of Artificial Intelligence, Beijing Key Laboratory of Multimedia and Intelligent Software Technology, Faculty of Information Technology, Beijing University of Technology, Beijing 100124, China, and also with the Beijing Research Institute, UBTECH, Beijing 100089, China.

Shaofan Wang, Dehui Kong, Lichun Wang, and Baocai Yin are with the Beijing Institute of Artificial Intelligence, Beijing Key Laboratory of Multimedia and Intelligent Software Technology, Faculty of Information Technology, Beijing University of Technology, Beijing 100124, China (e-mail: kdh@bjut.edu.cn).

Color versions of one or more figures in this article are available at https://doi.org/10.1109/TCYB.2021.3100507.

Digital Object Identifier 10.1109/TCYB.2021.3100507

video surveillance, motion retrieval, sports video analysis, and healthcare [1]–[4]. With the development of depth sensors, such as Microsoft Kinect, Nintendo Wii, and ASUS Xtion Pro Live, depth information can be acquired expediently. In fact, the early work of Johansson [5] has suggested that human actions could be modeled by the motion of a set of skeleton joints. In particular, with the release of Microsoft Kinect SDK [6], by which 3-D human skeleton joints can be recovered in real time, it is reasonable to treat equivalently the action video recorded by the RGB-D sequences with the sequence of 3-D positions of human skeleton joints. Therefore, it has motivated recent research work to investigate 3-D human action recognition using the skeleton joints data from depth sensors, and a lot of advanced methods have been proposed during the past few years [7]–[13].

While fruitful research work is devoted to 3-D skeleton-based action-recognition methods [14]–[16], existing methods mainly suffer from three problems.

First, the action sequence admits a highly complicated articulated change with probably repeated motion fragments, while current work is difficult to deeply exploit. Actually, most of the 3-D skeleton-based methods extracted features by pairwise displacement of joint positions within the current frame or pairwise adjacent frames. While some of the methods [17]–[22] treat the displacements by stacking them together, which produce a great amount of computing cost, others [23], [24] are devoted to subtle feature pruning and lead to complicated classification models.

Second, action sequence admits a great amount of noise due to the acquisition process from noncontact devices (e.g., Kinect sensor and time-of-flight cameras). In fact, previous methods usually achieve 5%–15% difference of accuracy between noncontact devices-based datasets (e.g., the `MSRAction3D` dataset [8]) and contacting devices-based datasets (e.g., the `CMU MoCap Database`[1]). Although some of the research work [25]–[27] circumvents this issue by refining or denoising schemes, those works take a great computational burden and remain far from being satisfactory.

Third, action recognition requires real-time implementation during some particular applications regarding interaction systems, such as human–computer interaction, electronic entertainment, and smart house technologies. While a few methods [17], [28]–[30] consider such an issue, most of them

[1]The dataset is publicly available at http://mocap.cs.cmu.edu.

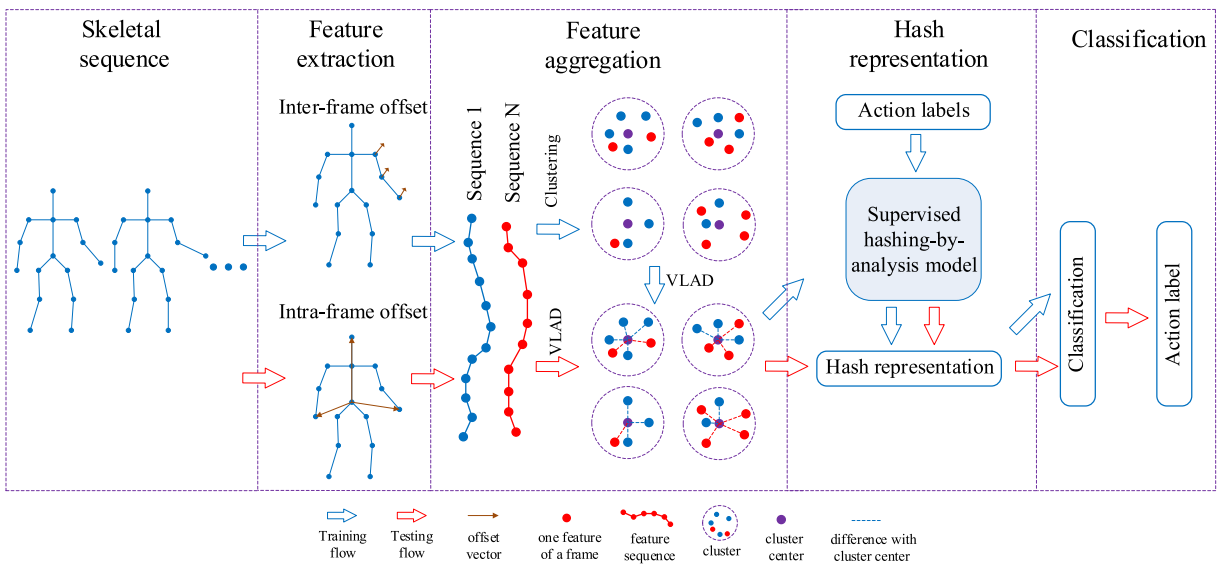

Fig. 1. Framework of the proposed method. For clarity, we only illustrate one interframe offset feature and one intraframe offset feature, and one $K$-means for one subset feature of joints.

are far from real-time implementation, which limits their wider applications.

To tackle the aforementioned problems, we propose a framework: LAKS + SHA for 3-D skeleton-based action recognition. The framework consists of two phases. In the low-level featuring phase, we propose a locally aggregated kinematic-guided skeletonlet (LAKS) to characterize each action sequence in an unsupervised fashion. LAKS is defined as the local aggregation of a few combinations of joint offsets grouped in terms of kinematic principle, which consists of the interframe and the intraframe offsets. In the high-level featuring phase, we propose a supervised hashing-by-analysis (SHA) model by integrating sparse representation with a hashing model. While the sparse representation learns an underlying "sparseland" of the high-dimensional feature represented by skeletonlet, the hashing representation maps the feature onto a compact Hamming space which facilitates an effective classification. A flowchart of LAKS + SHA is illustrated in Fig. 1.

The main contributions of our work are summarized as follows.

1) We propose the LAKS for representing actions, which exploits both the motion independence and the intrinsic relationship between body components. Besides, the representation effectively reduces the intraclass variation induced by noisy skeleton joints by considering the smooth movement of a sequence.

2) We propose an SHA model which combines the sparse representation theory with the hashing model. The resulting model improves the classification precision by learning the sparseland of the skeletonlet while maintaining high efficient computation by mapping the high-dimensional feature onto the Hamming space.

3) Our framework achieves both high accuracy and real-time implementation on three datasets of various

properties, one of which consists of noiseless actions, another of which admits a great number of noises, and the third of which contains human–object interactions.

The remainder of this article is organized as follows. Section II introduces the related work. Section III elaborates on action representation based on LAKS. Section IV proposes the SHA model and its optimization. We evaluate the performance of LAKS + SHA by the experimental results in Section V. The conclusion is given in Section VI.

## II. RELATED WORK

In this section, we briefly review research work related to our work. First, we review several skeleton-based human action-recognition methods from two aspects: 1) low-level feature extraction and 2) high-level feature representation. Then, we review hashing-based methods.

*Low-Level Feature Extraction Methods:* Many methods extracted features by pairwise displacement of joint positions within the current frame or between the current frame and the previous frame or first frame. Yang and Tian [19] proposed an action feature descriptor called eigenjoints by calculating the differences of joints, which includes relative position information, consecutive information, and offset information. Similarly, Jiang *et al.* [21] presented a method with consecutive information and relative position information. Lu *et al.* [22] extracted features by computing local position offsets of joints. Furthermore, some varietal feature extraction methods were proposed to tradeoff the accuracy and the expense of the features. Li and Leung [24] calculated the top-K relative variance of joint relative distance and determined which joint pairs should be selected. Cai *et al.* [29] calculated the position, speed, and acceleration of joints based on each limb. Chen *et al.* [23] proposed a part-based 5-D feature vector to explore the most relevant joints of body parts in each action and extracted the motion features by only the relevant joint positions. Luvizon *et al.* [30] proposed extracting sets of spatial and temporal local features from subgroups of joints, which can reduce the feature space. Yang *et al.* [31] exploited an architecture, called skelet, which is a combination of a subset of joints. Different from [31] which learns the class-specific skelet, we design a nonclass-specific skeletonlet according to the kinematic principle of human actions.

*High-Level Feature Representation Methods:* Some methods represented the actions using principal component analysis [19]; temporal pyramid [20], [32], [33]; graph model [23], [24]; probabilistic model [11], [34]; clustering [18], [22], [30]; etc. Wang *et al.* [32], [33] proposed a Fourier temporal pyramid to represent the temporal pattern and then introduced an actionlet ensemble representation to model the actions while capturing the intraclass variances. Luo *et al.* [20] proposed a temporal pyramid method combined with sparse coding and max-pooling function to represent the actions. Temporal pyramid-based methods are robust to noise and can characterize the temporal structure of the actions. However, the features have high dimensions and are redundant and time consuming for recognizing actions. Li and Leung [24] proposed a joint spatial graph to characterize the actions, in which the joint pairs with top-K relative distance variance were selected. Chen *et al.* [23] designed a score function for the action inference based on action graphs. Graph model-based methods are more robust to noise and insensitive to the length of sequences, but have high complexity. Xia *et al.* [11] proposed to model the temporal dynamics in action sequences with the hidden Markov model. Han *et al.* [34] presented a conditional random field to estimate the motion patterns in the manifold subspace. Probabilistic model-based methods can represent an action video in time order and model the dynamic patterns. However, the noisy joint position may undermine the performance. Zhu *et al.* [18] used bag of words (BOW) to represent the actions. They used $K$-means clustering to quantize the feature and constructed the histogram based on the cluster centers. Then, histogram features were extracted by counting the occurrences of the cluster centers. Luvizon *et al.* [30] used the vector of locally aggregated descriptors (VLADs) and a pool of clusters to represent an action and then proposed a metric-learning method which can efficiently combine the feature vectors. In general, clustering-based methods achieve small computational complexity. However, BOW has large quantization errors. Therefore, to have low computational complexity, and implement in real-time, we use VLAD to locally aggregate kinematic-guided skeletonlet.

*Hashing-Based Methods:* Hashing can encode documents, images, and videos by a set of binary codes, while preserving the similarity of the original data. Therefore, hashing has been successfully applied in many applications, such as information retrieval [35]–[37] and visual tracking [38]. Indyk and Motwani [39] introduced an approximate similarity search scheme based on locality-sensitive hashing. Wang *et al.* [36] proposed a semisupervised hashing method by minimizing empirical error on the labeled data while maximizing variance and independence of hash codes over the labeled and unlabeled data. Li *et al.* [37] proposed a hashing-across-Euclidean-space-and-Riemannian-manifold model for action recognition. The model embedded the two heterogeneous spaces into reproducing kernel Hilbert spaces, respectively, and then transformed to the final common Hamming space. Shen *et al.* [35] proposed a supervised hashing framework, which formulated a joint learning objective that integrates hash code generation and linear classifier training. In summary, previous hashing-based classification and retrieval methods maintain acceptable accuracies while greatly improve computational efficiency. Nevertheless, our findings demonstrate that integrating hash coding with the sparse model of skeletonlet not only improves the computational efficiency but enhances the characterization of the underlying structure of actions.

## III. Locally Aggregated Kinematic-Guided Skeletonlet

In this section, we propose the LAKS as a compact representation of an action sequence. The main idea is to first define a skeletonlet, guided by human kinematics, as a few combinations of joint offsets, which consist of the interframe and the intraframe offsets. And then a codebook is obtained by the cluster centers of the offset feature set. Finally, LAKS is obtained by aggregating all the vectors of difference between the offset and the codebook into a high-dimensional vector.

### A. Notations and Preliminaries

Denote scalars, vectors, matrices, and sets by nonbold letters, bold lowercase letters, bold uppercase letters, and calligraphic uppercase letters, respectively. Denote $\mathbf{I}$ to be the identity matrix whose size is determined in context. Denote $\|\cdot\|$ to be the $\ell_2$ norm of a vector or the Frobenius norm of a matrix, and denote $\|\mathbf{A}\|_{2,1} = \sum_{i=1}^{m} \|[\mathbf{A}]_{i,:}\|$ to be the $\ell_{2,1}$ norm of a matrix $\mathbf{A} \in \mathbb{R}^{m \times n}$. We denote $\mathrm{sgn}(\cdot)$ to be the sign function which returns 1 if the input is positive and $-1$ otherwise. For vectors $\mathbf{v}_1, \ldots, \mathbf{v}_n \in \mathbb{R}^m$, denote

$$\mathrm{cat}(\mathbf{v}_1, \ldots, \mathbf{v}_n) : \underbrace{\mathbb{R}^m \times \cdots \times \mathbb{R}^m}_{n} \to \mathbb{R}^{mn} \qquad (1)$$

to be the concatenate operator, that is

$$\mathrm{cat}(\mathbf{v}_1, \ldots, \mathbf{v}_n) = \begin{bmatrix} \mathbf{v}_1^\top, \ldots, \mathbf{v}_n^\top \end{bmatrix}^\top. \qquad (2)$$

Let $\{(\mathcal{X}_n, c_n)\}_{n=1}^{N}$ be a training set of $N$ action sequences, each pair of which consists of an action label $c_n \in \{1, \ldots, C\}$ and a collection $\mathcal{X}_n$ of 3-D positions of all joints of all frames, that is

$$\mathcal{X}_n := \bigcup_{i=1}^{15} \bigcup_{j=1}^{J_n} \underbrace{\left( x_{i,j}^n, y_{i,j}^n, z_{i,j}^n \right)^\top}_{\mathbf{x}_{i,j}^n} \subseteq \mathbb{R}^3 \qquad (3)$$

where $C$ is the total number of classes of actions, $i$, $j$, and $n$ are, respectively, the index of the joint number (Fig. 2(a) shows a 15-skeleton human body model), the frame number, and the training data serial number, $\mathbf{x}_{i,j}^n$ denotes the 3-D position of the $i$th skeleton of the $j$th frame, and $J_n$ denotes the total frame number of the $n$th training data. The task of 3-D action recognition, for an input action sequence $\mathcal{X}$ of $J$ frames, is to predict its action label $c \in \{1, \ldots, C\}$.

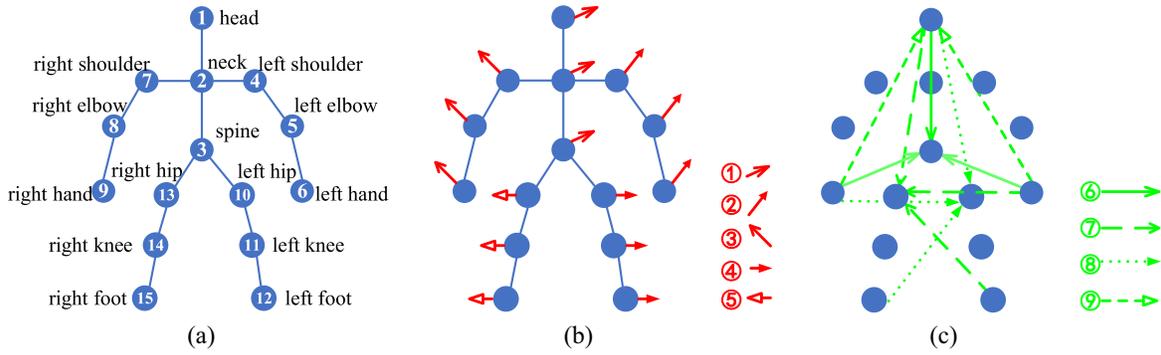

Fig. 2. Description of nine offset features of a human body model. (a) Fifteen-skeleton human body model. (b) Five interframe offset features. (c) Four intraframe offset features.

## B. Framewise Kinematic-Guided Skeletonlet

A skeletonlet is a mid-level granularity which indicates an underlying probable structure for characterizing different actions. Different from [31] which learns the class-specific *skelets*, we learn the nonclass-specific LAKS in an unsupervised fashion; the ingredients of LAKS are offset features defined by elaborately selecting a few combinations of skeleton joints according to the kinematic principle of human actions. Compared with the skelet [31], LAKS leverages the balance between the classification precision and computational efficiency.

The $\tau$-interval interframe offset of the $i$th joint of the $j$th frame for the $n$th training data is given by

$$\delta_{i,j}^n := \mathbf{x}_{i,j}^n - \mathbf{x}_{i,j-\tau}^n \tag{4}$$

where $\tau \in \mathbb{Z}^+$ denotes the time difference which balances the precision of the offset and the robustness to noise (i.e., the greater $\tau$ is, the more robust noise fluctuations are, and the lower the computation precision becomes; and vice versa). The intraframe offset of the $i$th, $i'$th skeletons of the $j$th frame for the $n$th training data is given by

$$\delta_{i,i':j}^n := \mathbf{x}_{i,j}^n - \mathbf{x}_{i',j}^n. \tag{5}$$

The framewise kinematic-guided skeletonlet, which reveals the framewise "visual words" property of an action sequence, consists of the interframe offset features and the intraframe offset features. Typically, the interframe offset features are selected by accumulating offsets of three interframe skeleton joints belonging to the same body components, which leads to five offset features of $\mathbb{R}^9$; the intraframe offset features are selected by accumulating intraframe offsets from a basic skeleton joint to its probably farthest joints, which leads to three offset features of $\mathbb{R}^9$ and an offset feature of $\mathbb{R}^6$. Fig. 2(b) and Table I: ①–⑤ demonstrate the construction of the interframe offset features, while Fig. 2(c) and Table I: ⑥–⑨ demonstrate the construction of the intraframe offset features. The nine offset features are denoted as follows:

$$\boldsymbol{\Delta}_{1,j}^n = \text{cat}\left(\delta_{1,j}^n, \delta_{2,j}^n, \delta_{3,j}^n\right)$$
$$\boldsymbol{\Delta}_{2,j}^n = \text{cat}\left(\delta_{4,j}^n, \delta_{5,j}^n, \delta_{6,j}^n\right)$$
$$\boldsymbol{\Delta}_{3,j}^n = \text{cat}\left(\delta_{7,j}^n, \delta_{8,j}^n, \delta_{9,j}^n\right)$$

### TABLE I
### Nine Offset Features

| Order | Type | Basic joint | Offset joints |
|---|---|---|---|
| ① | inter-frame | – | head, neck, spine |
| ② | inter-frame | – | left hand, left elbow, left shoulder |
| ③ | inter-frame | – | right hand, right elbow, right shoulder |
| ④ | inter-frame | – | left hip, left knee, left foot |
| ⑤ | inter-frame | – | right hip, right knee, right foot |
| ⑥ | intra-frame | spine | head, left hand, right hand |
| ⑦ | intra-frame | right hip | head, left hand, left foot |
| ⑧ | intra-frame | left hip | head, right hand, right foot |
| ⑨ | intra-frame | head | left hand, right hand |

$$\boldsymbol{\Delta}_{4,j}^n = \text{cat}\left(\delta_{10,j}^n, \delta_{11,j}^n, \delta_{12,j}^n\right)$$
$$\boldsymbol{\Delta}_{5,j}^n = \text{cat}\left(\delta_{13,j}^n, \delta_{14,j}^n, \delta_{15,j}^n\right)$$
$$\boldsymbol{\Delta}_{6,j}^n = \text{cat}\left(\delta_{3,1:j}^n, \delta_{3,6:j}^n, \delta_{3,9:j}^n\right)$$
$$\boldsymbol{\Delta}_{7,j}^n = \text{cat}\left(\delta_{13,1:j}^n, \delta_{13,6:j}^n, \delta_{13,12:j}^n\right)$$
$$\boldsymbol{\Delta}_{8,j}^n = \text{cat}\left(\delta_{10,1:j}^n, \delta_{10,9:j}^n, \delta_{10,15:j}^n\right)$$
$$\boldsymbol{\Delta}_{9,j}^n = \text{cat}\left(\delta_{1,6:j}^n, \delta_{1,9:j}^n\right) \tag{6}$$

with $\boldsymbol{\Delta}_{s,j}^n \in \mathbb{R}^9$, $1 \le s \le 8$, and $\boldsymbol{\Delta}_{9,j}^n \in \mathbb{R}^6$. The $n$th action sequence is then represented by

$$\left\{ \mathcal{S}_s^n := \bigcup_{j=1}^{J_n} \boldsymbol{\Delta}_{s,j}^n \right\}_{s=1}^9. \tag{7}$$

## C. Locally Aggregated Framewise Kinematic-Guided Skeletonlet

Motivated by the idea of the VLADs [40], we learn a codebook of $K$ visual words for each offset feature. That is, we assemble each of the nine offset features for all training data together, that is

$$\mathcal{S}_s := \bigcup_{n=1}^N \mathcal{S}_s^n = \bigcup_{n=1}^N \bigcup_{j=1}^{J_n} \boldsymbol{\Delta}_{s,j}^n, \ \ s = 1, \dots, 9 \tag{8}$$

and utilize the $K$-means clustering on $\mathcal{S}_s$ to yield $K$ subsets $\{\mathcal{S}_{s,k}^{(t)}\}_{k=1}^K$ and the corresponding centers $\{\mathbf{c}_{s,k}^{(t)}\}_{k=1}^K$,

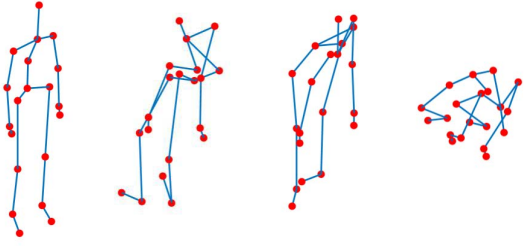

Fig. 3. Examples of a noiseless skeleton and three noisy skeletons from the `MSRAction3D` dataset.

that is

$$\left[\mathcal{S}_{s,1}^{(t)}, \ldots, \mathcal{S}_{s,K}^{(t)}\right] = K\text{means}(\mathcal{S}_s)$$
$$\mathbf{c}_{s,k}^{(t)} = \frac{1}{\left|\mathcal{S}_{s,k}^{(t)}\right|} \sum_{\mathbf{\Delta}' \in \mathcal{S}_{s,k}^{(t)}} \mathbf{\Delta}' \qquad (9)$$

$t = 1, \ldots, T, s = 1, \ldots, 9$, where the subscript $t$ (T, resp.) denotes the time (total times, resp.) of $K$-means clustering, as the $K$-means clustering yields different results for different initial values.

Before obtaining LAKS, it is necessary to handle noisy action frames as the noise data lower the description power of LAKS. Fig. 3 shows a noiseless action frame and three noisy action frames induced by the noise of capturing process. It is problematic to utilize these noisy joint positions directly to represent human skeletons. To handle this problem, we observe that an action sequence admits a smooth movement and tends to maintain a small change within adjacent frames. In other words, noiseless frames tend to form a cluster of greater cardinality while noisy frames tend to form a cluster of smaller cardinality. Motivated by this observation, we consider a cluster as a noisy cluster if its cardinality is less than a predefined threshold $\epsilon$, and replace all the features within it by the features of the corresponding previous frame.

Thus, a component of LAKS for the $n$th training data is given by summing the difference between an offset feature and its nearest voxel word (i.e., the cluster center) over all the offset features associated with the $n$th training data. That is

$$\overline{\mathbf{\Delta}}_{s,k,t}^n = \sum_{\mathbf{\Delta} \in \mathcal{S}_{s,k}^{(t)} \bigcap \mathcal{S}_s^n} \left(\mathbf{\Delta} - \mathbf{c}_{s,k}^{(t)}\right). \qquad (10)$$

LAKS of the $n$th training data is finally represented by concatenating the vector set $\cup_{s=1}^9 \cup_{k=1}^K \cup_{t=1}^T \overline{\mathbf{\Delta}}_{s,k,t}^n$ to form the following vector of $\mathbb{R}^{78KT}$:

$$\mathbf{y}_n = \text{cat}\left(\bigcup_{s=1}^9 \bigcup_{k=1}^K \bigcup_{t=1}^T \overline{\mathbf{\Delta}}_{s,k,t}^n\right). \qquad (11)$$

## IV. Supervised Hashing-by-Analysis Model

To better extract discriminant information and capture the intraclass variance, we propose an SHA model in this section as a high-level representation of human actions, which is effective and efficient for complicated action recognition.

### A. Learning Supervised Hashing-by-Analysis Model

Now that we have a compact representation $\{\mathbf{y}_n\}_{n=1}^N$ of the training set, we apply the power law normalization [30] to each $\mathbf{y}_n$, that is

$$\mathbf{y}_n \leftarrow \text{sgn}(\mathbf{y}_n)|\mathbf{y}_n|^{\frac{1}{2}}, \ n = 1, \ldots, N \qquad (12)$$

where both $\text{sgn}(\cdot)$ and $|\cdot|$ are elementwise operators, and form them as a matrix

$$\mathbf{Y} = [\mathbf{y}_1, \mathbf{y}_2, \ldots, \mathbf{y}_N]. \qquad (13)$$

Our goal is to learn the hash codes of all the training samples, that is, to learn the projection matrix $\mathbf{W} \in \mathbb{R}^{L \times 78KT}$ such that $\mathbf{B} = [\mathbf{b}_1, \ldots, \mathbf{b}_N] \in \{-1, 1\}^{L \times N}$, where $\mathbf{b}_i = \text{sgn}(\mathbf{W}^\top \mathbf{y}_i)$, and $L$ is the hash code length in the Hamming space.

Motivated by Shen *et al.* [35], which proposes a supervised hashing for image classification, we formulate the supervised hashing model for action recognition as follows:

$$\min_{\mathbf{W}, \mathbf{Q}, \mathbf{B}} \ \left\|\mathbf{L} - \mathbf{Q}^\top \mathbf{B}\right\|_F^2 + \lambda_1 \|\mathbf{Q}\|_F^2$$
$$\text{s.t.} \ \ \mathbf{B} = \text{sgn}\left(\mathbf{W}^\top \mathbf{Y}\right). \qquad (14)$$

The first term represents the classification error, where $\mathbf{L} = [\mathbf{l}_1, \ldots, \mathbf{l}_N] \in \mathbb{R}^{C \times N}$ is the ground-truth label matrix of training data, $C$ is the class number, $\mathbf{l}_n \in \mathbb{R}^C$ is the label vector of the $n$th data whose $c$th component takes one and other components take zero, and $\mathbf{Q}^\top \mathbf{B}$ is a linear predictive classifier. The second term represents the regularization on the classification matrix $\mathbf{Q} \in \mathbb{R}^{L \times C}$, and $\lambda_1 > 0$ is the regularization parameter.

Nevertheless, a gap between the compact representation $\mathbf{Y}$ of LAKS and the hashing projection mapping $\mathbf{W}^\top$ exists. The reason is that hash coding which is recognized as a discretization scheme in the sense of a set of cutting hyperplanes, might fail to characterize the compact feature space of LAKS due to its uniform distribution of energy. In contrast, sparse models tend to concentrate the energy of compact feature within very few components. As a result, it is expected that adopting hashing to supervised sparse coding of LAKS outperforms directly adopting hashing to LAKS. Based on the aforementioned argument, we utilize an analysis sparse constraint on $\mathbf{Y}$ and propose the following SHA model:

$$\min_{\mathbf{W}, \mathbf{Q}, \mathbf{B}, \mathbf{T}} \ \left\|\mathbf{L} - \mathbf{Q}^\top \mathbf{B}\right\|_F^2 + \lambda_1 \|\mathbf{Q}\|_F^2 + \lambda_2 \sum_{c=1}^C \left\|\mathbf{T}\mathbf{Y}_{[c]}\right\|_{2,1}$$
$$\text{s.t.} \ \ \mathbf{B} = \text{sgn}\left(\mathbf{W}^\top \mathbf{T}\mathbf{Y}\right) \qquad (15)$$

where $\mathbf{Y}_{[c]}$ denotes the training set of the $c$th class, $\mathbf{T} \in \mathbb{R}^{d \times 78KT}$ is an analysis dictionary which sparsifies the analysis representation of $\mathbf{Y}$, the third term adopts a group sparse constraint over the training data from the same class, $\mathbf{W} \in \mathbb{R}^{L \times d}$ is the projection matrix, and $\lambda_2 > 0$ is the regularization parameter.

## B. Optimization of SHA

Due to the discrete code $\mathbf{b}_i$, the optimization problem (15) is intractable. We relax (15) by replacing the hard constraint $\mathbf{B} = \text{sgn}(\mathbf{W}^\top \mathbf{TY})$ with a soft constraint, and introducing auxiliary variables $\mathbf{T}'_c \in \mathbb{R}^{d \times N_c}$ for replacing $\mathbf{TY}_{[c]}$, $c = 1, \ldots, C$, $N_c$ denotes the number of training data of the $c$th class, with $N = \sum_{c=1}^{C} N_c$. Model (15) is then transformed into

$$\min_{\mathbf{W}, \mathbf{Q}, \mathbf{B}, \mathbf{T}, \mathbf{T}'_c} \left\| \mathbf{L} - \mathbf{Q}^\top \mathbf{B} \right\|_F^2 + \lambda_1 \|\mathbf{Q}\|_F^2 + \lambda_3 \left\| \mathbf{B} - \mathbf{W}^\top \mathbf{TY} \right\|_F^2$$
$$+ \sum_{c=1}^{C} \left( \lambda_2 \|\mathbf{T}'_c\|_{2,1} + \frac{\mu}{2} \left\| \mathbf{T}'_c - \mathbf{TY}_{[c]} + \frac{1}{\mu} \mathbf{\Lambda}_c \right\|_F^2 \right)$$
$$\text{s.t. } \mathbf{B} \in \{-1, 1\}^{L \times N} \qquad (16)$$

where $\mathbf{\Lambda}_c \in \mathbb{R}^{d \times N_c}$ denotes the Lagrange multiplier matrix corresponding to the constraint $\mathbf{D}_c = \mathbf{DS}_{[c]}$, $c = 1, \ldots, C$, and $\lambda_1, \lambda_2, \lambda_3 > 0$ are regularization parameters. Model (16) is solved via the alternating direction method.

*The W-Subproblem:*

$$\min_{\mathbf{W}} \left\| \mathbf{B} - \mathbf{W}^\top \mathbf{TY} \right\|_F^2 \qquad (17)$$

which is a least-squares problem. According to the first-order optimization condition, the $\mathbf{W}$-subproblem has the following closed-form solution:

$$\mathbf{W} = \left( \mathbf{TYY}^\top \mathbf{T}^\top \right)^{-1} \mathbf{TYB}^\top. \qquad (18)$$

*The Q-Subproblem:*

$$\min_{\mathbf{Q}} \left\| \mathbf{L} - \mathbf{Q}^\top \mathbf{B} \right\|_F^2 + \lambda_1 \|\mathbf{Q}\|_F^2 \qquad (19)$$

which is a regularized least-squares problem. Again, the $\mathbf{Q}$-subproblem has the following closed-form solution:

$$\mathbf{Q} = \left( \mathbf{BB}^\top + \lambda_1 \mathbf{I} \right)^{-1} \mathbf{BL}^\top. \qquad (20)$$

*The B-Subproblem:*

$$\min_{\mathbf{B} \in \{-1,1\}^{L \times N}} \left\| \mathbf{L} - \mathbf{Q}^\top \mathbf{B} \right\|_F^2 + \lambda_3 \left\| \mathbf{B} - \mathbf{W}^\top \mathbf{TY} \right\|_F^2 \qquad (21)$$

which is an NP-hard problem because of the hashing constraint. To solve the $\mathbf{B}$-subproblem, we use the discrete cyclic coordinate descent (DCC) method [35] and optimize $\mathbf{B}$ row by row. To be specific, we denote $\mathbf{O} = \mathbf{QL} + \lambda_3 \mathbf{W}^\top \mathbf{TY}$ and decompose (21) into a quadratic term and a linear term, that is

$$\min_{\mathbf{B} \in \{-1,1\}^{L \times N}} \left\| \mathbf{Q}^\top \mathbf{B} \right\|_F^2 - 2\text{tr}\left( \mathbf{B}^\top \mathbf{O} \right). \qquad (22)$$

Then, we denote $[\mathbf{A}]_{i,:}$ to be the $i$th row of a matrix $\mathbf{A}$, and denote $[\mathbf{A}]_{-i,:}$ to be the submatrix of $\mathbf{A}$ consisting all but the $i$th row of $\mathbf{A}$. When updating the $l$th row of $\mathbf{B}$, $1 \le l \le L$, we fix other rows and transform (22) into

$$\min_{[\mathbf{B}]_{l,:} \in \{-1,1\}^{1 \times N}} \left( [\mathbf{Q}]_{l,:} [\mathbf{Q}]_{-l,:}^\top [\mathbf{B}]_{-l,:} - [\mathbf{O}]_{l,:} \right) [\mathbf{B}]_{l,:}^\top \qquad (23)$$

which has the following closed-form solution:

$$[\mathbf{B}]_{l,:} = \text{sgn}\left( [\mathbf{O}]_{l,:} - [\mathbf{Q}]_{l,:} [\mathbf{Q}]_{-l,:}^\top [\mathbf{B}]_{-l,:} \right). \qquad (24)$$

*The T-Subproblem:*

$$\min_{\mathbf{T}} \lambda_3 \left\| \mathbf{B} - \mathbf{W}^\top \mathbf{TY} \right\|_F^2 + \sum_{c=1}^{C} \left( \frac{\mu}{2} \left\| \mathbf{T}'_c - \mathbf{TY}_{[c]} + \frac{1}{\mu} \mathbf{\Lambda}_c \right\|_F^2 \right) \qquad (25)$$

which has the following closed-form solution:

$$\mathbf{T} = \left( \mu \mathbf{I} + 2\lambda_3 \mathbf{WW}^\top \right)^{-1}$$
$$\cdot \left( \mu \sum_{c=1}^{C} \left( \left( \mathbf{T}'_c + \frac{\mathbf{\Lambda}}{\mu} \right) \mathbf{Y}_{[c]}^\top \right) + 2\lambda_3 \mathbf{WBY}^\top \right) \left( \mathbf{YY}^\top \right)^{-1}. \qquad (26)$$

*The $T'_c$-Subproblem:*

$$\min_{\mathbf{T}'_c} \lambda_2 \|\mathbf{T}'_c\|_{2,1} + \frac{\mu}{2} \left\| \mathbf{T}'_c - \mathbf{TY}_{[c]} + \frac{1}{\mu} \mathbf{\Lambda}_c \right\|_F^2 \qquad (27)$$

which has the following closed-form solution:

$$[\mathbf{T}'_c]_{i,:} = \frac{\max\left( \|[\mathbf{U}_c]_{i,:}\|_2 - \frac{\lambda_2}{\mu}, 0 \right)}{\|[\mathbf{U}_c]_{i,:}\|_2} [\mathbf{U}_c]_{i,:}, \ i = 1, \ldots, d \qquad (28)$$

with $\mathbf{U}_c = \mathbf{TY}_{[c]} - (1/\mu)\mathbf{\Lambda}_c$, $c = 1, \ldots, C$. Note that the coefficient $[(\max(\|[\mathbf{U}_c]_{i,:}\|_2 - (\lambda_2/\mu), 0))/(\|[\mathbf{U}_c]_{i,:}\|_2)]$ degenerates to zero if the denominator vanishes.

We give algorithms for training and testing LAKS + SHA in Algorithms 1 and 2, respectively.

## V. Experimental Results

In this section, we evaluate our method on three publicly available human action datasets: 1) the MSRAction3D dataset [8]; 2) the UTKinectAction3D dataset [11]; and 3) the Florence3DAction dataset [41]. We first introduce three datasets (Section V-A). Then, we conduct parameter analysis (Section V-B) and ablation study (Section V-C). Afterward, we compare LAKS + SHA with several state-of-the-art methods on three datasets (Section V-D). Finally, we show the effectiveness of LAKS + SHA by listing the phasewise testing time (Section V-E). The experiments run on a Core i7-4790 3.6-GHz machine with 8-GB RAM using MATLAB R2016a, and the parameters are listed in Table II. The value of time difference $\tau$ is selected according to [16] and [30], while the values of $K$-means time T and cluster number $K$ are selected according to [30]. The selection of the other parameters is detailed in Section V-B.

### A. Datasets

MSRAction3D consists of 567 action sequences and is performed by ten individuals, including 20 classes: 1) high wave; 2) horizontal arm wave; 3) hammer; 4) hand catch; 5) forward punch; 6) high throw; 7) draw X; 8) draw tick; 9) draw circle; 10) hand clap; 11) hand wave; 12) side boxing; 13) bend; 14) forward kick; 15) side kick; 16) jogging; 17) tennis swing; 18) tennis serve; 19) golf swing; and 20) pickup throw. The data are recorded with a depth sensor similar to Kinect. Each individual performs each action twice or three times. The actions are suitable for gesture control or

---

**Algorithm 1:** Training of LAKS + SHA

**input** : $\{(\mathcal{X}_n, c_n)\}_{n=1}^N$, $K$means parameters $K, T$, code length $L$, atom number $d$, noisy frame threshold $\epsilon$, maximum iteration number; parameters $\lambda_1, \lambda_2, \lambda_3, \mu_{max}, \rho$

**output** : $\{\mathbf{c}_{s,k}^{(t)}\}_{s=1,k=1,t=1}^{9,K,T}, \mathbf{W}, \mathbf{Q}, \mathbf{B}, \mathbf{T}$

// Frame-wise kinematic-guided skeletonlet

1 Compute nine offset features via (6);

2 Obtain the total kinematic-guided offset feature $\{\mathcal{S}_s^n\}_{s=1}^9$ for the $n$th training data via (7);

// Locally aggregated frame-wise kinematic-guided skeletonlet

3 **for** $s = 1, \dots, 9$ **do**

4     Apply $K$means to $\mathcal{S}_s$ via (9) with $T$ times;

5 **end**

6 **foreach** $\mathbf{\Delta}_{s,j}^n \in \mathcal{S}_{s,k}^{(t)}$ such that $|\mathcal{S}_{s,k}^{(t)}| < \epsilon$ **do**

7     $\mathbf{\Delta}_{s,j}^n \leftarrow \mathbf{\Delta}_{s,j-1}^n$;

8 **end**

9 **for** $n = 1, \dots, N$ **do**

10     Obtain the locally aggregated skeletonlet $\mathbf{s}_n$ for the $n$th training data via (11);

11 **end**

// Supervised hashing-by-analysis representation

**initialize:** $\mathbf{Q}, \mathbf{T}, \mathbf{T}_c' \sim$ **Uniform**$(0, 1)$, $\mathbf{B} \sim$ **Uniform**$\{0, 1\}$, $\mu \leftarrow 1$

12 Apply power law normalization to $\{\mathbf{y}_n\}_{n=1}^N$ via (12);

13 **while** Not Converge $\wedge$ Not Reach Maximum Iteration **do**

14     Solve $\mathbf{W}$-subproblem (17) via Eq. (18);

15     Solve $\mathbf{Q}$-subproblem (19) via Eq. (20);

    // Solve $\mathbf{B}$-subproblem (21)

16     **for** $l = 1, \dots, L$ **do**

17        Update the $l$th row of $\mathbf{B}$ via Eq. (24);

18     **end**

19     Solve $\mathbf{T}$-subproblem (25) via Eq. (26);

20     Solve $\mathbf{T}_c'$-subproblem (27) via Eq. (28), $c = 1, \dots, C$;

21     Update Lagrange multiplier matrix by $\mathbf{\Lambda}_c \leftarrow \mathbf{\Lambda}_c + \mu(\mathbf{T}_c' - \mathbf{TY}_{[c]})$, $c = 1, \dots, C$;

22     Update penalty parameter by $\mu \leftarrow \min(\rho\mu, \mu_{max})$;

23 **end**

---

**Algorithm 2:** Testing of LAKS + SHA

**input** : Action sequence $\mathcal{X}$ of $J$ frames, $\mathbf{W}, \mathbf{T}$, $\{\mathbf{c}_{s,k}^{(t)}\}_{s=1,k=1}^{9,K,T}$ of training set

**output** : Action label $c$

// Frame-wise kinematic-guided skeletonlet

1 Compute offset features $\{\mathbf{\Delta}_{s,j}^{\text{test}}\}_{s=1,j=1}^{9,J}$ of $\mathcal{X}$ via (6) ;

2 Obtain the total kinematic-guided offset feature $\left\{\mathbf{S}_s^{\text{test}} := \cup_{j=1}^J \mathbf{s}_{s,j}^{\text{test}}\right\}_{s=1}^9$ of $\mathcal{X}$ via (7);

// Locally aggregated frame-wise kinematic-guided skeletonlet

3 $\mathbf{y}_{\text{test}} = \text{cat}\left(\bigcup_{s=1}^9 \bigcup_{k=1}^K \bigcup_{t=1}^T \sum_{\mathbf{\Delta} \in \mathcal{S}_s^{\text{test}}} (\mathbf{\Delta} - \mathbf{c}_{s,k}^{(t)})\right)$;

4 Apply power law normalization to $\mathbf{y}_{\text{test}}$ via (12);

// Supervised hashing-by-analysis representation

5 Compute hashing code $\mathbf{b}_{\text{test}} = \text{sgn}(\mathbf{W}^\top \mathbf{T}\mathbf{y}_{\text{test}})$;

6 Obtain final label $c$ using nearest neighbor (NN) classifier of $\mathbf{b}_{\text{test}}$ on $\mathbf{B}$ with Hamming distance;

---

UTKinectAction3D consists of ten subjects, of which nine are males and one is female. One of the person is left-handed. The ten actions include: 1) walk; 2) sit down; 3) stand up; 4) pick up; 5) carry; 6) throw; 7) push; 8) pull; 9) wave; and 10) clap hands. Each subject performs each action twice in various views, and the length of actions ranges from 5 to 120 frames. The 3-D positions of 15 joints are available in each frame.

Florence3DAction consists of nine activities: 1) wave; 2) drink from a bottle; 3) answer phone; 4) clap; 5) tight lace; 6) sit down; 7) stand up; 8) read watch; and 9) bow, each of which is captured by a Kinect sensor. During the acquisition, ten subjects were asked to perform the above actions. The 3-D positions of 15 joints are available in each frame. Moreover, most activities involve human–object interactions, which is challenging for recognition only by 3-D joints.

### B. Parameter Analysis

To evaluate the sensitivity of our proposed method to parameters, we show details for parameter selection, including noisy frame threshold $\epsilon$, the number of atoms $d$, the code length $L$, and other regularization parameters $\lambda_1, \lambda_2$, and $\lambda_3$. We set ranges of these six parameters empirically: $\lambda_1 \in [0.001, 10]$, $\lambda_2 \in [0.001, 10]$, $\lambda_3 \in [1e-5, 1e-2]$, $\epsilon \in [10, 70]$, $d, L \in [16, 128]$. Then, we empirically sample a few combinations of parameter values and obtain an initial optimal value combination. Afterward, we finetune each parameter over each range while fixing other parameters according to the initial value combination. Fig. 4(a)–(f) shows the effects of different parameters. For parameters $\lambda_1$ and $\lambda_3$, LAKS + SHA achieves a robust performance in a wide parameter range on three datasets, whose optimal ranges are $\lambda_1 \in [0.1, 1]$ and $\lambda_3 \in [1e-5, 1e-2]$, respectively. LAKS + SHA is insensitive to parameter $\lambda_2$ on the UTKinectAction3D dataset and is robust in the range $[0.1, 10]$ on the other two datasets. For noisy frame threshold $\epsilon$, we can observe that

---

TABLE II
PARAMETER SETTING IN OUR EXPERIMENTS

| Parameter | Interpretation | Value |
|-----------|----------------|-------|
| $\lambda_1$ | regularization parameter | 1 |
| $\lambda_2$ | regularization parameter | 1 |
| $\lambda_3$ | regularization parameter | $1e-3$ |
| $\epsilon$ | noisy frame threshold | 30, 20, 20§ |
| $L$ | code length | 32 |
| $d$ | atom number | 64 |
| $\tau$ | time difference | 2 |
| $T$ | K-means time | 5 |
| $K$ | cluster number | 23 |

§Correspond to MSRA, UTKinect, Florence

---

electronic gaming. The dataset provides 3-D skeleton joints and depth maps. We use 3-D skeleton joints only, and the 3-D coordinates of 20 joints are available in each frame.

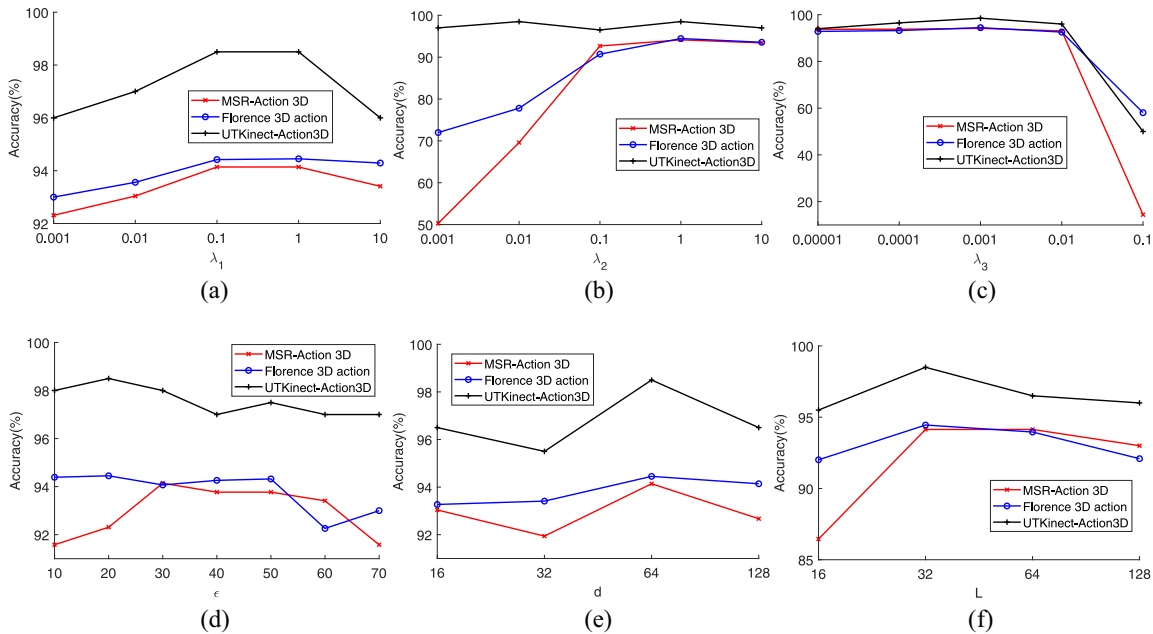

Fig. 4. Performance variations of LAKS + SHA with various values of parameters for three datasets. (a) Regularization parameter $\lambda_1$. (b) Regularization parameter $\lambda_2$. (c) Regularization parameter $\lambda_3$. (d) Noisy frame threshold $\epsilon$. (e) Atom number $d$. (f) Code length $L$.

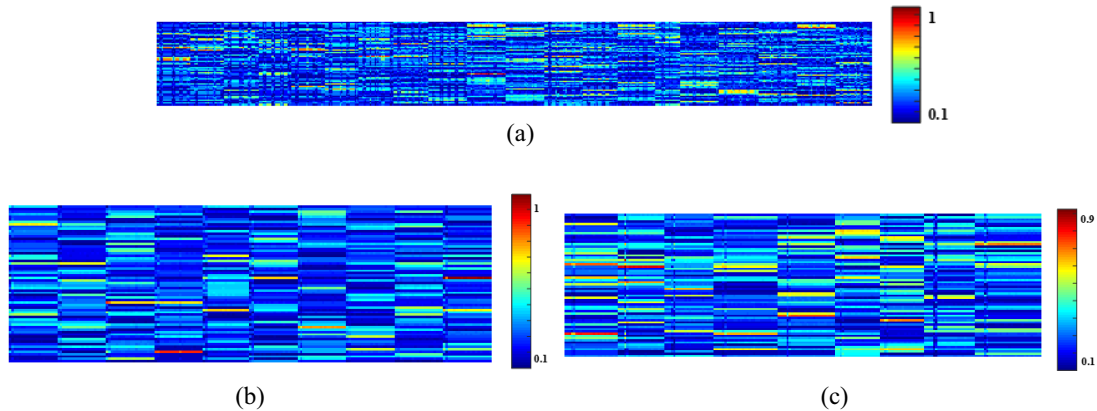

Fig. 5. Magnitude of the sparse coding **TY** of (15) for three datasets. The $x$-axis represents the samples arranged by action classes, and the $y$-axis represents the coding. (a) MSRAction3D. (b) UTKinectAction3D. (c) Florence3DAction.

LAKS + SHA reaches the better performance when $\epsilon$ takes 30, 20, and 20 for MSRAction3D, UTKinectAction3D, and Florence3DAction, respectively. This result shows that when $\epsilon$ is set to the above values, we can remove frames that are not conducive to the performance of LAKS + SHA. If $\epsilon$ is too large, noiseless frames may be removed, which may affect the performance of LAKS + SHA. If $\epsilon$ is too small, the noisy frames may not be completely removed, which also may affect the performance of LAKS + SHA. Especially for the MSRAction3D dataset which has more noisy frames, using noisy frame threshold $\epsilon$ significantly improves the performance of LAKS + SHA, which can be seen from Fig. 4(d). For the number of atoms $d$ and the code lengths $L$, considering the performance and efficiency of LAKS + SHA, we can observe that LAKS + SHA can reach the better performance when $d = 64$ and $L = 32$. Table II shows the best parameters setting of each dataset.

### C. Ablation Study

In this section, we conduct ablation experiments to evaluate the contribution of different components of LAKS + SHA. As LAKS + SHA mainly consists of two stages: 1) low-level feature representation (Sections III-B and III-C) and 2) high-level feature representation (Section IV), we conduct the ablation experiments from these two aspects.

We conduct the first ablation study with six low-level feature representation methods [9], [18]–[20], [29], [30] for action recognition in the upper part of Table III, each of which is followed by a common classifier: NN. According to the results, LAKS generally performs better than other features.

Next, we conduct the second ablation study to evaluate the effectiveness of high-level feature representation, that is, the SHA model (15). We compare LAKS + SHA with three models: LAKS, LAKS + SA, and LAKS + SH with common low-level feature: LAKS, and show the results in the lower part of Table III. The result shows that LAKS + SHA achieves significant improvement. In fact, LAKS + SHA which adopts hashing to supervised sparse coding of LAKS outperforms LAKS + SH which directly adopts hashing to LAKS. It means that sparse models which concentrate the energy of compact feature within very fewer components can serve as a bridge between LAKS and hashing coding. Fig. 5 shows the magnitudes of sparse coding with the group sparse constraint in LAKS + SHA on three datasets. The horizontal coordinate represents the samples, and the vertical coordinate represents the dimensionality. From the figures, we can see that using group sparse constraint from the same class, the intraclass variations among features are compressed. That is, the codes of the same dimension among samples maintain almost consistency in the same class.

TABLE III
ABLATION STUDY OF LOW-LEVEL AND HIGH-LEVEL FEATURE REPRESENTATIONS ON THREE DATASETS

| Model | Low-level feature | High-level feature | MSRA | UTKinect | Florence |
|---|---|---|---|---|---|
| Luo *et al.* [20] | center-symmetric motion local ternary pattern | - | 79.04 | - | - |
| Yang *et al.* [19] | eigenjoints | - | 82.3 | - | - |
| Cai *et al.* [29] | active skeleton representation with $K$-means | - | 83.38 | - | - |
| Ohn-bar *et al.* [9] | joint angles similiarities | - | 83.53 | - | - |
| Zhu *et al.* [18] | spatiotemporal joints features | - | - | 87.9 | - |
| Luvizon *et al.* [30] | VLAD of spatial and temporal local features | - | 83.15 | **90.97** | 90.71 |
| LAKS | locally aggregated kinematic-guided skeletonlet (11) | - | **85.71** | 89.47 | **91.82** |
| LAKS | | - | 85.71 | 89.47 | 91.82 |
| LAKS+SA | | (15) without hashing coding **B** | 91.58 | 97.5 | 92.5 |
| LAKS+SH | locally aggregated kinematic-guided skeletonlet (11) | (15) without analysis dictionary **T** | 91.21 | 98 | 91.29 |
| LAKS+SHA | | (15) | **94.14** | **98.5** | **94.45** |

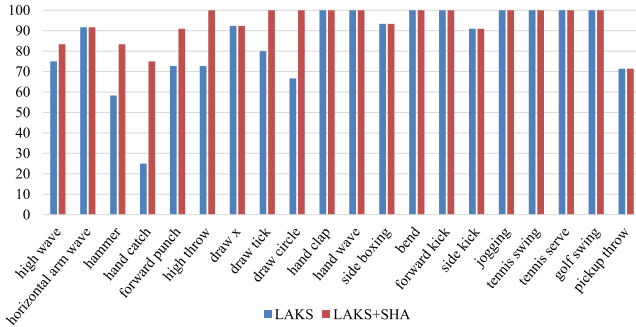

Fig. 6. Classwise accuracy of LAKS versus LAKS + SHA on MSRAction3D.

Finally, let us further examine how LAKS + SHA outperforms LAKS. Fig. 6 shows the classwise accuracy of LAKS and LAKS + SHA on MSRAction3D. We observe that for those highly similar actions (e.g., **hand catch** and **hammer**), LAKS + SHA can significantly improve the accuracy. To qualitatively investigate feature representations of LAKS and LAKS + SHA, we utilize t-SNE visualization on feature representations of LAKS [i.e., **Y** of (15)] and LAKS + SHA [i.e., **B** of (15)]. Fig. 7(a) shows the visualization of feature representations of 20 classes of MSRAction3D. Compared with LAKS, LAKS + SHA can further reduce the distance of intraclass and increase the distance of interclasses. Moreover, Fig. 7(b) specifically visualizes feature representations of the first ten classes, which are mainly hand movements and have similar actions. We observe that LAKS + SHA distinguishes similar actions clearly. In contrast, LAKS confuses a few samples of different classes and produces small interclass distance, which easily leads to misclassification.

TABLE IV
QUANTITATIVE RESULTS (ACCURACY) ON THREE DATASETS

| Datasets | Methods | Accuracy(%) |
|---|---|---|
| MSRAction3D | Li *et al.* [8] | 74.7 |
| | Xia *et al.* [11] | 79 |
| | Yang *et al.* [19] | 82.3 |
| | Oreifej *et al.* [26] | 85.8 |
| | Cai *et al.* [29] | 91.01 |
| | Wang *et al.* [43] | 91.40 |
| | Zanfir *et al.*[28] | 91.7 |
| | Devanne *et al.* [13] | 92.1 |
| | Li *et al.* [24] | 92.2 |
| | Luo *et al.* [20] | 93.8 |
| | **LAKS+SHA** | **94.14** |
| UTKinectAction3D | Slama *et al.* [44] | 88.5 |
| | Xia *et al.* [11] | 90.5 |
| | Devanne *et al.* [13] | 91.5 |
| | Wang *et al.* [25] | 93.5 |
| | Chen *et al.* [33] | 85.96 |
| | Wang *et al.* [43] | 96.48 |
| | Luvizon *et al.* [30] | 98 |
| | **LAKS+SHA** | **98.5** |
| Florence3DAction | Seidenari *et al.* [42] | 82.15 |
| | Devanne *et al.* [13] | 87.04 |
| | Anirudh *et al.* [45] | 89.67 |
| | Vemulapalli *et al.* [46] | 90.88 |
| | Li *et al.* [24] | 93.2 |
| | Wang *et al.* [43] | 94.25 |
| | **LAKS+SHA** | **94.45** |

Based on those projected 2-D t-SNE features, we further conduct a significance test using two independent samples t-test. Fig. 8 shows the *p*-value of each pair of classes, where lower triangular values denote *p*-values of the first dimension of t-SNE features, and upper triangular values denote *p*-values of the second dimension of t-SNE features. The *p*-value denotes the probability of observing the given result by chance if the null hypothesis is true. Small values of *p* cast doubt on the validity of the null hypothesis. That is to say, the greater the value of *p* is, the smaller the significant difference between the two classes is; and vice versa. As we can see from Fig. 8, LAKS + SHA distinguishes many pairs of classes significantly, which further verifies the discriminative power of LAKS + SHA.

### D. Comparative Results on Three Datasets

To compare with other state-of-the-art methods, we evaluate LAKS + SHA on MSRAction3D. We use the cross-subject test setting described in [26], that is, half of the action classes of which are utilized for training, and others are for testing. Table IV shows the accuracy comparison of different methods, which are reported in the corresponding paper. LAKS + SHA can obtain a recognition accuracy of 94.14%, which is better than all other methods. This is a very good performance considering the unstable skeleton joints and the huge intraclass variance among different individuals. Fig. 9(a) shows the confusion matrix of action recognition corresponding to the best accuracy by LAKS + SHA. The vertical coordinate represents the true label of action, and the horizontal coordinate represents the recognition result. The result shows that the recognition accuracies of most actions reach or approach 100% except for a few highly similar actions, such as **hand catch** and **hammer**. These action pairs share the most similar active skeleton.

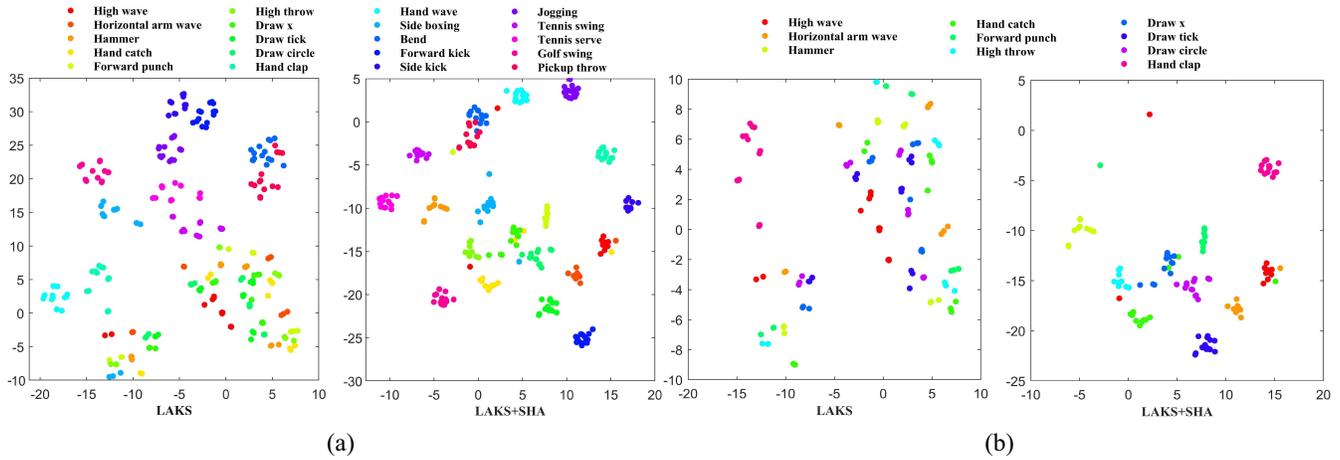

Fig. 7. T-SNE visualization of feature representations of LAKS and LAKS + SHA on `MSRAction3D`. (a) T-SNE visualization of 20 classes of LAKS and LAKS + SHA. (b) T-SNE visualization of ten classes of LAKS and LAKS + SHA.

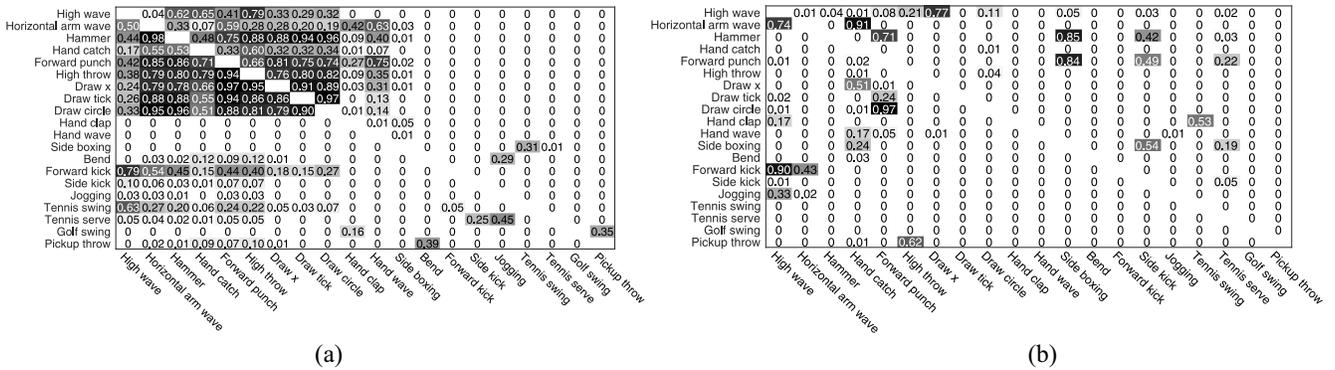

Fig. 8. T-test results of LAKS and LAKS + SHA on `MSRAction3D`. Lower triangular values denote *p*-values of the first dimension of t-SNE features, while upper triangular values denote *p*-values of the second dimension of t-SNE features. (a) T-test result of LAKS. (b) T-test result of LAKS + SHA.

We also evaluate LAKS + SHA on `UTKinectAction3D`. We follow the standard leave-one-out-cross-validation protocol in [11] to evaluate LAKS + SHA on this dataset. Table IV shows the results for LAKS + SHA compared with the state-of-the-art methods. Here, we list the accuracies of these methods which are reported in the corresponding paper with the same way of evaluation. From Table IV, we can see that the average recognition accuracy of LAKS + SHA is 98.5% which is sufficiently high accuracy. Fig. 9(b) shows the confusion matrix of action recognition of LAKS + SHA. From this figure, we can see that the recognition accuracy of each action is at least 95%.

For `Florence3DAction`, we follow the standard leave-one-out-cross-validation protocol [41] to evaluate LAKS + SHA. Table IV shows the accuracy comparison of different methods, which are reported in the corresponding paper. According to Table IV, LAKS + SHA achieves the best performance (94.45%) among all methods. Fig. 9(c) shows the confusion matrix of action recognition. The result shows that the recognition accuracies of most actions reach or approach 100% except for **drink from a bottle**, **answer phone**, and **read watch**. The reason is that these actions involve human–object interactions, which are unavailable from the skeletons.

### E. Efficiency Analysis

Table V shows the average testing time of each phase of our method for each action on the three datasets, where the testing time includes four phases: 1) feature extraction; 2) feature aggregation; 3) hash representation; and 4) classification. For all the three datasets, LAKS + SHA only

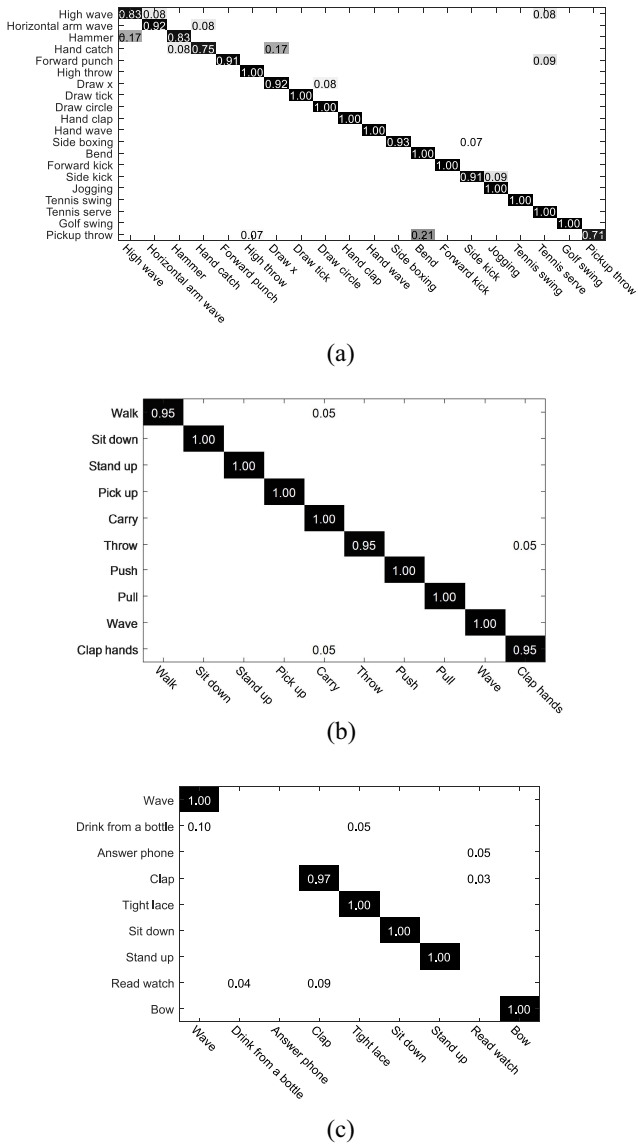

Fig. 9. Confusion matrices of LAKS + SHA on three datasets. The vertical coordinate represents the true label of an action, and the horizontal coordinate represents the recognition result. (a) MSRAction3D. (b) UTKinectAction3D. (c) Florence3DAction.

#### TABLE V
PHASEWISE COMPUTATIONAL COMPLEXITY AND RUNTIME (UNIT: MS) OF TESTING LAKS + SHA AVERAGED BY ACTION SEQUENCE

| Phase | Line num of Algorithm 2 | Complexity | MSRA | UTKinect | Florence |
|---|---|---|---|---|---|
| Feature extraction | lines 1-2 | $\mathcal{O}(J_n)$ | 2.103 | 1.597 | 1.031 |
| Feature aggregation | lines 2-4 | $\mathcal{O}(J_n)$ | 6.852 | 6.901 | 6.841 |
| Hash representation | line 5 | $\mathcal{O}(1)$ | 0.096 | 0.099 | 0.092 |
| Classification | line 6 | $\mathcal{O}(NJ_n)$ | 0.931 | 0.720 | 0.922 |
| Average testing time | – | – | 9.982 | 9.317 | 8.886 |

requires 9.982, 9.317, and 8.886 ms per sequence with the average sequence length of 39, 28, and 19 frames, respectively. In other words, LAKS + SHA produces remarkable performance in computational efficiency. For MSRAction3D, LAKS and LAKS + SHA produce the same feature extraction time (2.103 ms) and feature aggregation time (6.852 ms). For LAKS, the subsequent classification time is 12.945 ms.

However, for LAKS + SHA, the sum of subsequent hash representation time and classification time is only 1.027 ms. Since the dimension of LAKS is $78KT = 8970$, while the dimension of LAKS + SHA is only 32. Therefore, LAKS + SHA accelerates the classification time.

We also do a computational complexity analysis and show the result in Table V. The computational complexity of feature extraction, feature aggregation, and hash representation are exactly $9J$, $9TJ$, and $78KTL+L$, respectively. In our experiment, the number of times T, the cluster number $K$, and the code length $L$ are set as 5, 23, and 32, respectively. In a word, LAKS + SHA has low computational complexity, which can be implemented in real time.

## VI. CONCLUSION

In this article, we propose a novel framework for skeleton-based human action recognition. We design a skeletonlet which is guided by human kinematics as a few combinations of joint offsets, which consist of the interframe and the intraframe offsets. The feature extraction method can reduce the feature space and be better clustered and simultaneously consider the rationality of motion. Then, a locally aggregated framewise kinematic-guided skeletonlet is proposed to characterize the actions, which can better describe the distribution of the feature vectors with respect to the center compared with BOW. Besides, considering the smooth movement of the human action sequence, we process the irregular skeletons or the impossible motions caused by the noisy joint estimations, which can improve the robustness of the representation method. Furthermore, an SHA model is proposed, which combines the sparse representation theory with the hashing model. We use sparse vectors to represent the low-level feature, considering the group sparse making the samples consistent in the same class. Then, the hashing representation maps the feature onto a compact Hamming space which facilitates the effective classification. The experiments are performed on three 3-D human action datasets, including MSRAction3D, UTKinectAction3D, and Florence3DAction. The experimental results validate that the proposed method can be implemented in real time and achieves a higher recognition accuracy compared with many state-of-the-art methods. In future work, we will use other features to enrich the representation model and investigate the open problem of recognizing multiview actions.

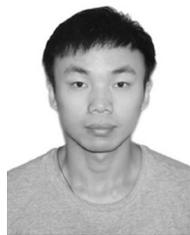

**Bin Sun** received the Ph.D. degree from the Beijing Key Laboratory of Multimedia and Intelligent Software Technology, Faculty of Information Technology, Beijing University of Technology, Beijing, China, in 2019.

His research interests include pattern recognition and machine learning.

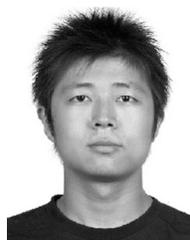

**Shaofan Wang** (Member, IEEE) received the B.S. and Ph.D. degrees in computational mathematics from the Dalian University of Technology, Dalian, China, in 2003 and 2010, respectively.

He is an Associate Professor with the Beijing Key Laboratory of Multimedia and Intelligent Software Technology, Faculty of Information Technology, Beijing University of Technology, Beijing, China. His research interests include pattern recognition and machine learning.

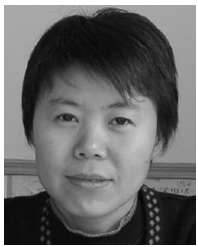

**Dehui Kong** (Member, IEEE) received the M.S. and Ph.D. degrees from Beihang University, Beijing, China, in 1992 and 1996, respectively.

She is currently a Professor with the Beijing Key Laboratory of Multimedia and Intelligent Software Technology, Faculty of Information Technology, Beijing University of Technology, Beijing. Her research interests include virtual reality, computer graphics, and pattern recognition.

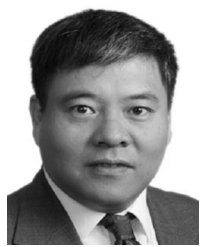

**Baocai Yin** received the B.S., M.S., and Ph.D. degrees from the Dalian University of Technology, Dalian, China, in 1985, 1988, and 1993, respectively.

He is a Professor and the Dean of the Beijing Key Laboratory of Multimedia and Intelligent Software Technology, Faculty of Information Technology, Beijing University of Technology, Beijing, China. His research interests include digital multimedia, multifunctional perception, virtual reality, and computer graphics.

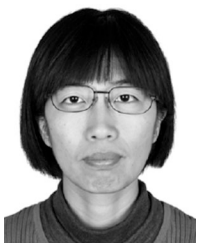

**Lichun Wang** received the M.S. degree from the Harbin Institute of Technology, Harbin, China, in 1998, and the Ph.D. degree from Nanjing University, Nanjing, China, in 2001.

She is a Professor with the Beijing Key Laboratory of Multimedia and Intelligent Software Technology, Faculty of Information Technology, Beijing University of Technology, Beijing, China. Her research interests are human–computer interaction and machine learning.